\documentclass{Interspeech}

\interspeechcameraready

\usepackage[T5]{fontenc}
\usepackage[utf8]{inputenc}
\usepackage{xcolor} 
\usepackage{tabularray}
\usepackage[most]{tcolorbox}
\usepackage{float}
\newtcolorbox[auto counter]{problem}[1][]{%
    enhanced,
    breakable,
    colback=white,
    colbacktitle=white,
    coltitle=black,
    fonttitle=\bfseries,
    boxrule=.6pt,
    titlerule=.2pt,
    toptitle=3pt,
    bottomtitle=3pt,
    title=Problem~\thetcbcounter,
    #1}

\definecolor{SoftGreen}{rgb}{0.0, 0.5, 0.0}

\title{\textsc{ViToSA}: Audio-Based Toxic Spans Detection on\\Vietnamese Speech Utterances}

\author[affiliation={1,3}]{Huy}{Ba Do}
\author[affiliation={2,3}]{Vy}{Le-Phuong Huynh}
\author[affiliation={2,3}]{Luan}{Thanh Nguyen}

\affiliation{Faculty of Computer Science}{University of Information Technology}{Ho Chi Minh City, Vietnam}
\affiliation{Faculty of Information Science and Engineering}{University of Information Technology}{\\Ho Chi Minh City, Vietnam}
\affiliation{}{Vietnam National University Ho Chi Minh City}{Vietnam}

\email{\{21522137,20520951\}@gm.uit.edu.vn, luannt@uit.edu.vn}

\usepackage{comment}

\begin{document}

\maketitle

\begin{abstract}
    
    Toxic speech on online platforms is a growing concern, impacting user experience and online safety. While text-based toxicity detection is well-studied, audio-based approaches remain underexplored, especially for low-resource languages like Vietnamese. This paper introduces \textsc{ViToSA} (\textbf{Vi}etnamese \textbf{To}xic \textbf{S}pans \textbf{A}udio), the first dataset for toxic spans detection in Vietnamese speech, comprising 11,000 audio samples (25 hours) with accurate human-annotated transcripts. We propose a pipeline that combines ASR and toxic spans detection for fine-grained identification of toxic content. Our experiments show that fine-tuning ASR models on \textsc{ViToSA} significantly reduces WER when transcribing toxic speech, while the text-based toxic spans detection (TSD) models outperform existing baselines. These findings establish a novel benchmark for Vietnamese audio-based toxic spans detection, paving the way for future research in speech content moderation\footnote{\url{https://github.com/vitosa-research/ViToSA-Dataset}}.

\textbf{\textcolor{red}{Disclaimer:}} This paper includes real examples from social media platforms that may be perceived as toxic or offensive.

\end{abstract}

\section{Introduction and Related Work}

In the context of robust digital content development, online platforms have become increasingly popular for community interaction and information sharing; however, the rise of toxic audio utterances has become a significant concern \cite{Namin2018TheSO,yousefi2021audio,10.1007/978-3-319-15168-7_26}. Furthermore, the widespread dissemination of sensitive and toxic phrases and audio clips is having a negative impact on users' mental well-being as well as on individual honor and dignity \cite{article,bucur-etal-2021-exploratory,Volkow2021}. Such content can incite violence, promote hatred, and inflict deep psychological harm on listeners, especially children and teenagers who are particularly vulnerable. The uncontrolled spread of toxic speech in online audio environments not only degrades communication quality but also creates a negative atmosphere, making many users feel concerned and even withdraw from discussions, as mentioned in the work of Qayyum et al. \cite{10386402}. This growing issue undermines public trust in digital platforms and threatens a safe, healthy communication environment.

Research on detecting toxic speech in audio has gained attention, but it remains relatively underdeveloped compared to text-based approaches. Efforts such as the DeToxy dataset \cite{ghosh22b_interspeech} and MuTox \cite{costa-jussa-etal-2024-mutox} have introduced toxic speech datasets along with classification models, including end-to-end and multilingual approaches. Besides, Nada et al. \cite{nada2023lightweight}'s studies have explored efficient models for real-time toxicity detection, while Liu et al. \cite{liu24h_interspeech} have investigated the integration of speech and text modalities to improve detection accuracy. However, existing studies often focus on utterance-level classification and lack the ability to detect toxic segments within speech.

\begin{figure}[t!]
    \centering
    \includegraphics[width=\linewidth]{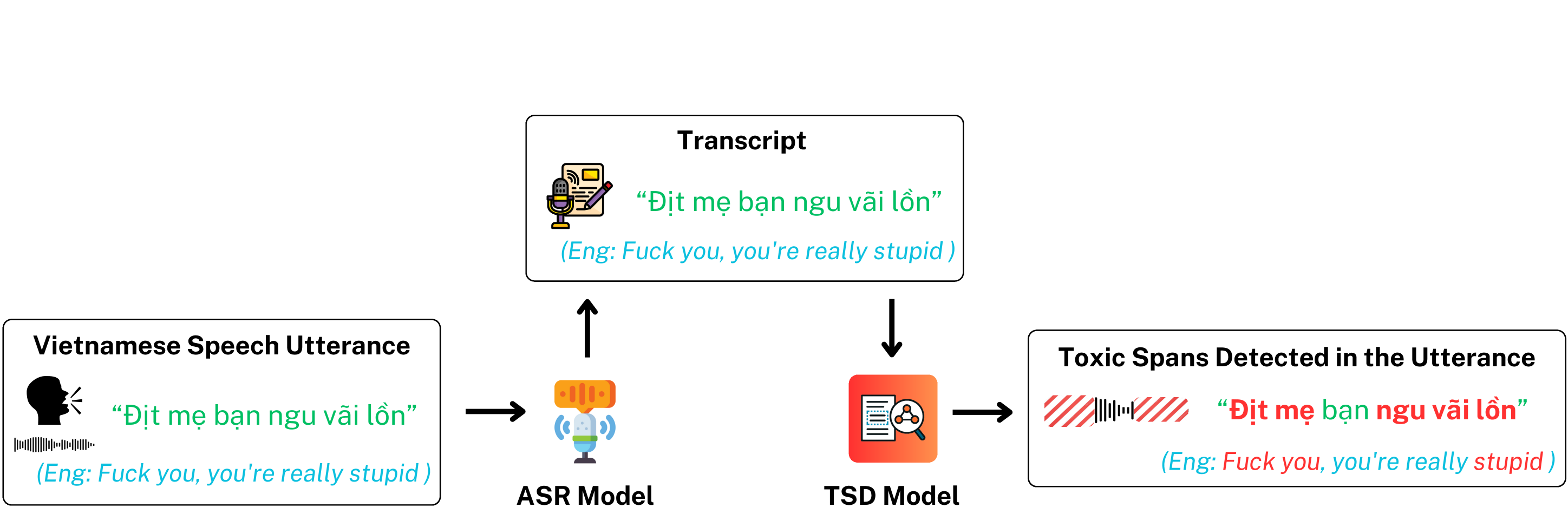}
    \caption{Framework of \textsc{ViToSA}.}
    \label{fig:frame_work}
\end{figure}

This issue affects many languages but is particularly severe in under-resourced contexts like Vietnamese, which lacks sufficient datasets and where research on toxic content detection has largely focused on text. Notable contributions include UIT-ViCTSD \cite{Nguyen_2021} and ViHSD \cite{luu2021large}, which aim to identify toxicity and hate speech in user comments; the ViHOS dataset \cite{hoang-etal-2023-vihos}, designed for detecting toxic textual phrases; and the ViHateT5 model \cite{thanh-nguyen-2024-vihatet5}, which utilizes a text-to-text transformer for various hate-speech-related tasks in Vietnamese. While these studies have advanced text-based toxicity detection, there is currently no dedicated dataset or research specifically addressing toxic speech in Vietnamese audio. This gap highlights the urgent need for developing resources and methodologies to detect and mitigate toxic speech in Vietnamese, ensuring a more comprehensive approach to online safety.

In this research, we aim to address the limitations of existing studies and meet practical needs by developing a comprehensive approach for detecting toxic speech in Vietnamese audio. To address these challenges, our contributions are as follows: (1) We introduce the novel \textsc{ViToSA} dataset, the first specifically focused on Vietnamese toxic audio segments, comprising 25 hours of speech, along with a dedicated evaluation test set serving as a benchmark for both ASR and TSD tasks; (2) We propose an effective pipeline that integrates a domain-specific speech recognition model with a text-based language model to accurately detect toxic audio segments; (3) We highlight the impact of our approach, demonstrating its potential to advance toxic speech detection in low-resource languages and lay the groundwork for future research in this field.
\section{\textsc{ViToSA} Dataset}

We begin by conducting preliminary experiments to evaluate the performance of existing ASR and TSD models in Vietnamese. These experiments are designed to assess the effectiveness of current models in transcribing toxic speech, identify specific challenges faced by ASR systems when handling toxic content, and evaluate the accuracy of TSD models in identifying toxic spans. Both tasks are evaluated using our \textsc{ViToSA} test set, which was constructed prior to the training set to serve as a reliable benchmark. The test set includes triplets of audio, manually transcribed text, and annotated toxic spans, providing a comprehensive resource for analyzing model performance.

\subsection{Preliminary Experiments}
\label{sec:preliminary}

\textbf{ASR Task.} Recent state-of-the-art ASR models for Vietnamese, such as Whisper \cite{whisper_2023}, Wav2Vec2 (W2V2)\footnote{Vietnamese variants of the W2V2 architecture include wav2vec2-base-vi-vlsp2020 and wav2vec2-base-vietnamese-250h, available on HuggingFace.} \cite{baevski2020wav2vec}, and PhoWhisper \cite{PhoWhisper}, demonstrate strong performance on standard benchmarks and widely used speech corpora. However, these models are typically trained on datasets with limited coverage of toxic vocabulary, leading to subpar performance in recognizing toxic speech, as shown in Table \ref{tab:preliminary}.

\begin{table}[ht]
\centering
\caption{Outputs of four models on a Vietnamese toxic utterance.}
\label{tab:preliminary}
\resizebox{\linewidth}{!}{%
\begin{tblr}{
  cell{1}{1} = {r=3}{},
  hline{1,8} = {-}{0.08em},
  hline{4} = {-}{dashed},
}
\textbf{Ground Truth} & thì \textcolor{red}{\textbf{đéo}} nói \textcolor{red}{\textbf{địt mẹ mày}} nó đến bốn năm lần thì đỡ thế \textcolor{red}{\textbf{lồn}} nào được\\
 & \textit{(Eng: hadn't \textcolor{red}{\textbf{fucking}} said anything, but \textcolor{red}{\textbf{your fucking mother~}}came four}\\
 & \textit{or five times, how the~\textbf{\textbf{\textcolor{red}{fuck}}}~could it be better?)}\\
Whisper & để theo nỗi bịt mày à mày nói đến bốn năm lần đi đó thế là một nào được\\
wav2vec2-base-vi-vlsp2020 & thì đơ nói \textcolor{red}{\textbf{địt mày}} mày nói đến bung năm lần thì nói thế lần nào được\\
wav2vec2-base-vietnamese-250h & vì đi nói đi mày á nói đến bốn năm lần thì nói thế lần nào được\\
PhoWhisper & vì \textbf{\textcolor{red}{đéo}} nói mấy ảnh mày nói đến bốn năm lần vì đỡ thế lộn vào đường.
\end{tblr}
}
\end{table}

Due to the frequent misrecognition of toxic words in ASR outputs, we propose constructing a Vietnamese ASR dataset specifically for the toxic speech domain to improve model robustness in handling such content. 

\textbf{TSD Task.} 
We utilize the ViHOS dataset \cite{hoang-etal-2023-vihos}, which has been annotated to identify toxic text spans in Vietnamese. This dataset is used to fine-tune language models, optimizing their performance for the toxic spans detection task. After training, we evaluate these models on our test set to assess their effectiveness in detecting toxic content.

\subsection{Data Creation}
\begin{figure*}[t]
    \centering
    \includegraphics[width=\linewidth]{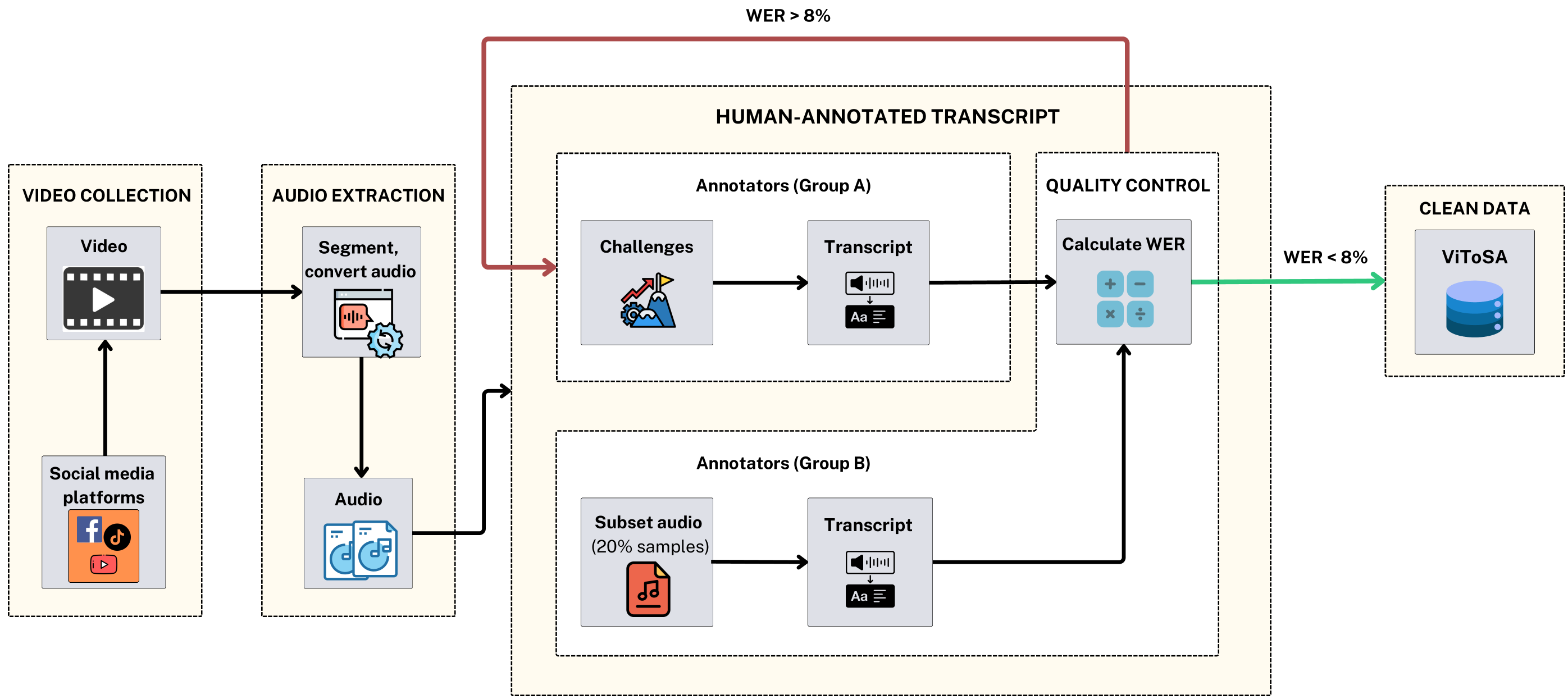}
    \caption{Pipeline for Collecting, Processing, and Quality Checking Transcribed Audio for the \textsc{ViToSA} dataset (train and validation).}
    \label{fig:data_process}
\end{figure*}

In this paper, we first release \textsc{ViToSA}, a high-quality dataset for Vietnamese speech processing, with a focus on toxic speech research. We strictly follow the process shown in Figure \ref{fig:data_process} to collect and annotate audio data. 


\textbf{Data Collection.} The Data Collection Process begins with the Video Collection phase. Short video clips containing toxic content are manually gathered from social media platforms such as Facebook, YouTube, and TikTok.

\textbf{Audio Extraction.} Once videos are collected, we extract audio files using the open-source library SoundFile\footnote{\url{https://python-soundfile.readthedocs.io/en/0.13.1/}} to convert videos into audio files. We hired undergraduate students from various academic backgrounds as annotators\footnote{Paid according to the local minimum wage.}, training them to identify toxic content and use the Audio Cutter\footnote{\url{https://mp3cut.net/}} tool for annotation. The selected audio segments range from 1 to 14 seconds in length. We discard segments shorter than 1 second due to a lack of meaningful context for toxicity identification, while those longer than 14 seconds are split to prevent cognitive overload for annotators.

\textbf{Human-annotated Transcription Phase} involves a challenge designed to help annotators understand the guidelines and improve transcript quality. We randomly select 50 audio files from the dataset for assessment. Annotators in group A are paired into sets $A_i = \{A_i \mid i \in N\}$, with each pair consisting of two members. Each annotator independently listens and transcribes the audio. The transcripts from each pair are then compared to calculate WER. The challenge consists of three rounds:

\begin{itemize}
    \item If a pair's WER is less than 8\%, they are considered to have met the standard and understood the guideline.
    \item If WER is greater than 8\%, we analyze the errors, update the guideline for clarification, and proceed to the next round.
\end{itemize}

After three rounds, all pairs achieved WER below 8\%, ensuring high-quality data. Once all annotators in group A met the standard, they proceeded to transcribe the remaining samples. To prevent fraud, members within the same pair were unaware of each other's identities, avoiding the risk of one member transcribing while the other merely copied.

\textbf{Quality Control Phase} is conducted by annotators in group B. 20\% of the samples from each annotator in group A are randomly selected and assigned to group B as ground truth. We continue using the WER threshold of 8\% to evaluate transcripts. If the WER between group A and group B is below 8\%, the transcript from that group A annotator is deemed valid. Otherwise, annotators exceeding the threshold must re-transcribe the entire set.

The final dataset contains 24.75 hours of Vietnamese-speaking utterances across 11,802 audio-transcript pairs\footnote{\url{https://huggingface.co/datasets/ViToSAResearch/ViToSA_Dataset}}, split into training, validation, and test sets, with 1,000 samples for testing and the rest divided 8:2 for training and validation.

\section{Methodology}

Having established the dataset, we now introduce our proposed detection framework, \textsc{ViToSA}, for detecting toxic speech segments in Vietnamese utterances. As shown in Figure \ref{fig:frame_work}, it consists of two key components: Automatic Speech Recognition (ASR), which transcribes spoken utterances, and Toxic Spans Detection (TSD), which identifies toxic segments in the transcriptions.

\subsection{Automatic Speech Recognition}
We utilize state-of-the-art transformer-based pre-trained models specifically optimized for Vietnamese automatic speech recognition. Trained on large-scale multilingual and monolingual corpora, these models effectively transcribe spoken language into text while demonstrating robustness to variations in accent, background noise, and speech patterns.

\begin{figure}[H]
    \centering
    \includegraphics[width=\linewidth]{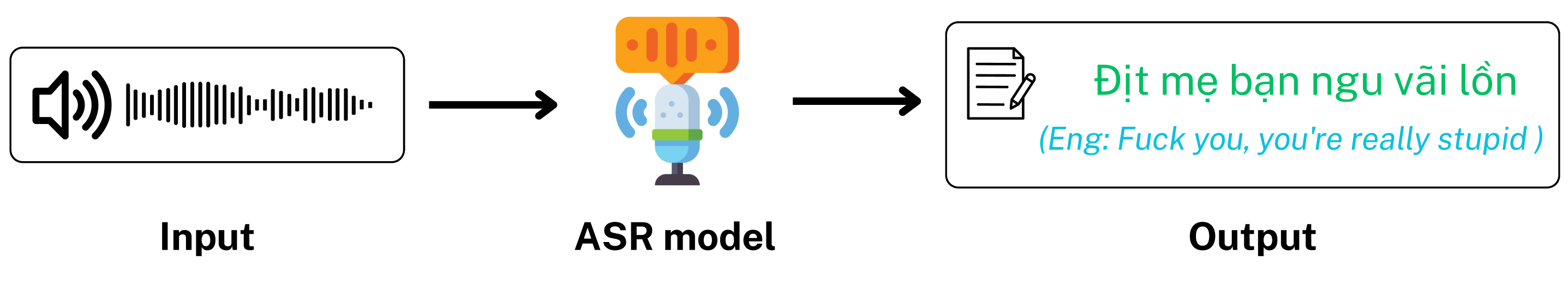}
    \caption{Input and output of the ASR component.}
    \label{fig:ASR_pipeline}
\end{figure}

To further improve their performance in recognizing toxic speech, we fine-tune these models on our domain-specific ASR dataset, \textsc{ViToSA}, enabling more accurate transcription of Vietnamese utterances containing toxic content.

\subsection{Toxic Spans Detection}
After obtaining transcriptions from the ASR component, we apply BERT-based language models, either Vietnamese-specific or multilingual, to detect and precisely localize toxic words or phrases within the text.

\begin{figure}[H]
    \centering
    \includegraphics[width=\linewidth]{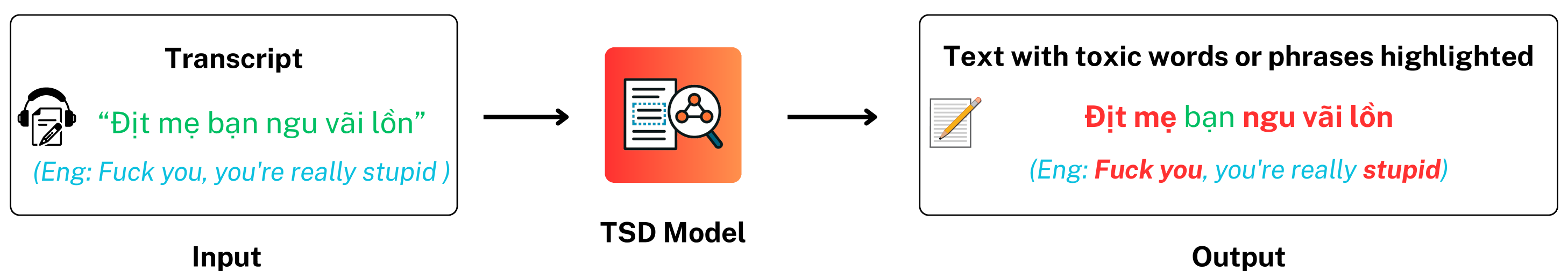}
    \caption{Illustration of the input and output of the TSD component. Predicted text-based spans are highlighted in bold red.}
    \label{fig:ASR_pipeline}
\end{figure}

By leveraging deep contextual embeddings, these models effectively capture semantic subtleties and syntactic patterns, enabling accurate identification of both explicit and context-dependent toxic language.
\section{Experiments}

We perform experiments on the \textsc{ViToSA} dataset, focusing on two key tasks: ASR and TSD. The process is outlined in the following sections: data pre-processing, evaluation metrics, and speech recognition experimental results.

\subsection{Data}

We use our \textsc{ViToSA} dataset to perform ASR.  All audio files are resampled to 16kHz and converted to mono channel to ensure consistency. For the ASR task, text pre-processing includes lowercasing, removing punctuation, and converting numbers into words, enhancing model readability and accuracy. In the TSD task, unnecessary whitespace is eliminated, line breaks are standardized, and toxic word positions are formatted into a structured labeling scheme, optimizing the data for precise identification.

For TSD, we use ViHOS training data for fine-tuning (as explained in Section \ref{sec:preliminary}). Those fine-tuned models are then evaluated on our \textsc{ViToSA} test set.

\subsection{Models and Settings}

We present the models and experimental settings used in our research on two tasks. Note that we use a single NVIDIA A100 GPU for all experiments in this study.

\textbf{ASR Models.}
For the ASR task, we utilize several models, including wav2vec2 \cite{baevski2020wav2vec} variants fine-tuned for Vietnamese, namely wav2vec2-base-vi-vlsp2020\footnote{\url{https://huggingface.co/nguyenvulebinh/wav2vec2-base-vi-vlsp2020}} and wav2vec2-base-vietnamese250h\footnote{\url{https://huggingface.co/nguyenvulebinh/wav2vec2-base-vietnamese-250h}}. Additionally, we employ the multilingual Whisper (base) model \cite{whisper_2023} and its Vietnamese fine-tuned version, PhoWhisper (base) \cite{PhoWhisper}. All ASR models are trained for 10 epochs with a batch size of 8, using the AdamW optimizer with a learning rate of 5e-5, a warmup ratio of 0.1, and a weight decay of 0.05.

\textbf{TSD Models.} We fine-tune current high-performance models in Vietnamese for TSD tasks, including multilingual pre-trained models such as XLM-R (base) \cite{conneau-etal-2020-unsupervised}, BERT (base, multilingual, cased) \cite{devlin-etal-2019-bert}, DistilBERT (base, multilingual, cased) \cite{sanh2019distilbert}, and monolingual ones such as PhoBERT (base, v2) \cite{nguyen-tuan-nguyen-2020-phobert}, ViSoBERT \cite{nguyen-etal-2023-visobert}, CafeBERT \cite{do-etal-2024-vlue}, and ViHateT5 \cite{thanh-nguyen-2024-vihatet5}. Note that ViHateT5 is already fine-tuned on the TSD task, we only use the model in the original paper without further fine-tuning. They are then fine-tuned for 4 epochs with a batch size of 8, using the AdamW optimizer. The training process is conducted with a learning rate of 2e-5 and a warmup ratio of 0.1 to optimize model performance.

\subsection{Evaluation Metrics}  
Word Error Rate (WER) is a widely used metric for assessing the accuracy of speech recognition models. Moreover, following the methodology mentioned in the work of Hoang et al. \cite{hoang-etal-2023-vihos}, we evaluate the toxic spans detection task using Accuracy (Acc), Macro F1 (MF1), and Weighted F1 (WF1) scores.  

\subsection{Results and Discussions}
We fine-tune ASR models on our annotated training data \textsc{ViToSA} and evaluate them on the test set, depicted in Table \ref{tab:result_asr}. For TSD, we fine-tune language models on ViHOS and assess their performance on our \textsc{ViToSA} test set, listed in Table \ref{tab:result_vihos_span}.

\begin{table}
\centering
\caption{Speech recognition experimental results on \textsc{ViToSA} test set.}
\label{tab:result_asr}
\resizebox{\linewidth}{!}{%
\begin{tblr}{
  row{2} = {c},
  row{7} = {c},
  cell{2}{1} = {c=4}{},
  cell{7}{1} = {c=4}{},
  hline{1,12} = {-}{0.08em},
  hline{2,7} = {-}{0.05em},
  hline{3,8} = {-}{dashed},
}
\textbf{Models}                                & \textbf{Toxic}                                                 & \textbf{ Non-toxic}                                           & \textbf{All}                                                  \\
\textbf{\textit{w/o \textsc{ViToSA} dataset}}       &                                                                &                                                               &                                                               \\
Whisper                                        & 1.660                                                          & 0.593                                                         & 1.149                                                         \\
wav2vec2-base-vi-vlsp2020                      & 0.988                                                          & 0.984                                                         & 0.986                                                         \\
wav2vec2-base-vietnamese-250h                  & 0.997                                                          & 0.999                                                         & 0.998                                                         \\
PhoWhisper                                     & 0.615                                                          & 0.212                                                         & 0.418                                                         \\
\textbf{\textbf{\textit{with \textsc{ViToSA} dataset}}} &                                                                &                                                               &                                                               \\
Whisper                                        & 0.325~\textsubscript{\textcolor{SoftGreen} {$\downarrow$ 1.335}} & 0.264~\textsubscript{\textcolor{SoftGreen}{$\downarrow$ 0.329}} & 0.289~\textsubscript{\textcolor{SoftGreen}{$\downarrow$ 0.860}} \\
wav2vec2-base-vi-vlsp2020                      & 0.319~\textsubscript{\textcolor{SoftGreen}{$\downarrow$ 0.669}}  & 0.302~\textsubscript{\textcolor{SoftGreen}{$\downarrow$ 0.682}} & 0.310~\textsubscript{\textcolor{SoftGreen}{$\downarrow$ 0.676}} \\
wav2vec2-base-vietnamese-250h                  & 0.342~\textsubscript{\textcolor{SoftGreen}{$\downarrow$ 0.655}}  & 0.280~\textsubscript{\textcolor{SoftGreen}{$\downarrow$ 0.719}} & 0.311~\textsubscript{\textcolor{SoftGreen}{$\downarrow$ 0.687}} \\
\textbf{PhoWhisper}                            & 0.302~\textsubscript{\textcolor{SoftGreen}{$\downarrow$ 0.313}}  & 0.192~\textsubscript{\textcolor{SoftGreen}{$\downarrow$ 0.020}} & 0.257~\textsubscript{\textcolor{SoftGreen}{$\downarrow$ 0.161}} 
\end{tblr}
}
\end{table}

\begin{table}
\centering
\caption{Toxic spans detection experimental results on \textsc{ViToSA} test set.}
\label{tab:result_vihos_span}
\begin{tblr}{
  cells = {c},
  hline{1,9} = {-}{0.08em},
  hline{2} = {-}{0.05em},
  hline{8} = {-}{dashed},
}
\textbf{Models}  & \textbf{Acc}   & \textbf{WF1}   & \textbf{MF1}   \\
ViHateT5         & 0.765          & 0.785          & 0.500          \\
DistilBERT       & 0.937          & 0.934          & 0.732          \\
BERT             & 0.940          & 0.940          & 0.768          \\
XLM-R            & 0.940          & 0.943          & 0.790          \\
CafeBERT         & 0.927          & 0.932          & 0.807          \\
ViSoBERT         & 0.945          & 0.947          & 0.817          \\
\textbf{PhoBERT} & \textbf{0.951} & \textbf{0.955} & \textbf{0.837} 
\end{tblr}
\end{table}

\textbf{The need for a domain-specific toxic audio dataset.} Table~\ref{tab:result_asr} underscores the importance of using a dedicated toxic audio dataset, such as \textsc{ViToSA}, for training ASR models. Without \textsc{ViToSA}, all models exhibit significantly higher word error rates (WER), particularly for toxic speech, where errors are more pronounced. After fine-tuning with \textsc{ViToSA}, WER drops considerably across all models, with Whisper experiencing the most significant improvement (from 1.149 to 0.289 overall). The wav2vec2-based models also benefit from \textsc{ViToSA}, showing WER reductions of approximately 0.676 and 0.687. Even PhoWhisper, which initially had a lower WER, further improves. These results confirm that general ASR models struggle with toxic speech due to data scarcity, and incorporating a domain-specific dataset significantly enhances performance, making ASR more reliable in toxic speech recognition tasks.

\textbf{TSD on normalized text (from ASR models) achieves higher performance than direct evaluation on social-media texts of ViHOS.} The results in Table~\ref{tab:result_span} indicate that performing TSD on normalized text, generated by ASR models, yields a higher MF1 compared to direct evaluation on the ViHOS dataset\footnote{According to the best performance in the original paper \cite{hoang-etal-2023-vihos} that obtained 0.772 MF1 with PhoBERT (large).}. Among all models, PhoBERT achieves the highest performance with 0.837 MF1, demonstrating its effectiveness in detecting toxic spans. ViSoBERT follows closely, particularly excelling in MF1 (0.817), indicating better generalization to minority toxic spans. Other transformer-based models, such as XLM-R and BERT, also perform well, with MF1 scores above 0.75. However, ViHateT5, which is mainly pre-trained and fine-tuned on social media texts, lags behind, particularly in MF1 (0.500), suggesting difficulties in handling the normalized toxic language. The superior performance on normalized text suggests that ASR-generated outputs, after normalization, may simplify TSD for general transformer-based models. However, this normalization process appears to challenge domain-specific pre-trained models like ViHateT5, which have been trained exclusively on social media data.

\textbf{Result Analysis.} To further assess model performance, we conduct inference again on a representative toxic utterance from Table~\ref{sec:preliminary} using the trained models. The predictions show that after training on \textsc{ViToSA}, nearly all toxic words in the utterance are accurately detected, a significant improvement compared to the initial results before fine-tuning on \textsc{ViToSA}. This demonstrates the necessity of constructing a dedicated dataset for toxic word recognition in Vietnamese utterances. Although the WER of the models remains relatively high, leading to some inaccuracies in full-sentence ASR outputs, our primary focus in this task is toxic word detection, for which the results are well-aligned with our objectives.

\begin{table}
\centering
\caption{Outputs of four models on a Vietnamese toxic utterance after fine-tuning on \textsc{ViToSA} dataset.}
\label{tab:result_span}
\resizebox{\linewidth}{!}{%
\begin{tblr}{
  cell{1}{1} = {r=3}{},
  hline{1,8} = {-}{0.08em},
  hline{4} = {-}{dashed},
}
\textbf{Ground Truth} & thì \textcolor{red}{\textbf{đéo}} nói \textcolor{red}{\textbf{địt mẹ mày}} nó đến bốn năm lần thì đỡ thế \textcolor{red}{\textbf{lồn}} nào được\\
 & \textit{(Eng: hadn't~\textcolor{red}{\textbf{\textbf{fucking}}}~said anything, but~\textcolor{red}{\textbf{\textbf{your fucking mother~}}}came}\\
 & \textit{four or five times, how the~\textbf{\textbf{\textbf{\textbf{\textcolor{red}{fuck}}}}}~could it be better?)}\\
Whisper & thì \textcolor{red}{\textbf{đéo}} nói \textcolor{red}{\textbf{địt mẹ mày}} nói đến bốn năm lần thì \textcolor{red}{\textbf{đéo}} thế \textcolor{red}{\textbf{lồn}} nào được\\
wav2vec2-base-vi-vlsp2020 & \textcolor{red}{\textbf{địt đéo}} nói \textcolor{red}{\textbf{địt mẹ mày}} nói đến bốn năm lần thì đỡ thế \textcolor{red}{\textbf{lồn}} nào được\\
wav2vec2-base-vietnamese-250h & thì \textcolor{red}{\textbf{đéo}} nói \textcolor{red}{\textbf{địt mẹ mày}} nó đến bốn năm lần thì đỡ thế \textcolor{red}{\textbf{lồn}} nao đường\\
PhoWhisper & thì \textcolor{red}{\textbf{đéo}} nói \textcolor{red}{\textbf{địt mẹ mày}} nó đến bốn năm lần thì đỡ tế \textcolor{red}{\textbf{lồn}} nào được
\end{tblr}
}
\end{table}
\section{Conclusion}

This paper introduces \textsc{ViToSA}, the first benchmark for detecting toxic spans in Vietnamese speech, addressing the gap in audio-based toxicity detection for low-resource languages. Our findings highlight the limitations of current ASR models in accurately transcribing toxic speech and demonstrate that fine-tuning ASR on \textsc{ViToSA} significantly reduces WER for toxic content. Furthermore, our Transformer-based toxic span detection models achieve strong results, showing the effectiveness of ASR-transcribed text for toxicity detection. We hope \textsc{ViToSA} fosters further research in speech-based toxicity detection and supports the development of a safer online environment.


\section{Acknowledgement}
This research was supported by The VNUHCM-University of Information Technology's Scientific Research Support Fund.

\bibliographystyle{IEEEtran}
\bibliography{mybib}

\end{document}